# Constructing Extreme Learning Machines with zero Spectral Bias

Kaumudi Joshi, Vukka Snigdha, Arya Kumar Bhattacharya*
Mahindra University, Hyderabad, India
arya.bhattacharya@mahindrauniversity.edu.in

*Abstract*—The phenomena of Spectral Bias, where the higher frequency components of a function being learnt in a feedforward Artificial Neural Network (ANN) are seen to converge more slowly than the lower frequencies, is observed ubiquitously across ANNs. This has created technology challenges in fields where resolution of higher frequencies is crucial, like in Physics Informed Neural Networks (PINNs). Extreme Learning Machines (ELMs) that obviate an iterative solution process which provides the theoretical basis of Spectral Bias (SB), should in principle be free of the same. This work verifies the reliability of this assumption, and shows that it is incorrect. However, the structure of ELMs makes them naturally amenable to implementation of variants of Fourier Feature Embeddings, which have been shown to mitigate SB in ANNs. This approach is implemented and verified to completely eliminate SB, thus bringing into feasibility the application of ELMs for practical problems like PINNs where resolution of higher frequencies is essential.

*Index Terms*—Artificial Neural Networks, Neural Tangent Kernel, Spectral Bias, Physics Informed Neural Networks, Extreme Learning Machines, Fourier Feature Embeddings.

## I. INTRODUCTION

It has been demonstrated that in the iterative process of training of a feedforward neural network, the lower frequency components of the function being learnt tend to converge much faster than the higher frequencies. This can be deduced from Neural Tangent Kernel (NTK) theory [1], where the error components related to the larger eigenvalues of the Kernel matrix, corresponding to the smaller (lower) frequencies, are shown to reduce much faster as compared to the smaller eigenvalues [2, 3]. This is known as the Spectral Bias in feedforward Artificial Neural Networks (ANNs).

A major area of development and application of ANNs is in the solution of coupled sets of differential equations that govern different domains in physics, like the Navier Stokes equations of fluid mechanics [4], the Maxwell's equations in electromagnetics [5], constitutive equations in structural mechanics, etc. They follow from the initial work of Karniadakis et al [6] and are generally known as Physics Informed Neural Networks (PINNs). PINNs have found significant application in diverse domains [7] and is a rapidly advancing field with multiple all-round applications. Investigations have shown [3] that the implicit spectral bias in ANNs apply more severely on the higher derivative terms of the differential equations being resolved through PINNs. There are many applications, as for example in modelling the smaller Kolmogorov scales of turbulence in fluid mechanics, where resolution of the higher frequency components become crucial – and the inherent spectral bias of ANNs amplified further into PINNs becomes a bottleneck in the evolution of the technology.

Extreme Learning Machines (ELMs) [8-10] have the baseline architecture of multilayer perceptrons with one hidden layer, and with the elements of the first weight matrix taken randomly. Their distinction from feedforward ANNs lies in the manner of training; the weights of the second weight matrix are solved directly by matrix inversion, instead of the iterative optimization process followed in ANNs. As a consequence, solution times are reduced by around two orders of magnitude, making them amenable to many real time operations specially where on-the-fly adaptive retraining is needed [10], where one would rule out ANNs. Because of their specific architectural style, they have not found adoption in the multifarious fields where Deep ANN variants continue to impact. Here it may be mentioned that most developments in PINNs use fully connected feedforward networks.

NTK Theory and concomitant Spectral Bias (SB) are fundamentally related to iterative convergence processes, and hence ELMs that are based on direct solutions should logically be free from SB. Consequently, they can in principle be used in PINNs (i.e. their variants) with ease of handling higher derivative terms and incorporating higher frequencies without encumbrance. But before trying to formulate ELMs for resolving sets of governing differential equations of physical systems, which would represent a fusion of two divergent technical streams, it is worth investigating if ELMs are actually free of any manifestations of SB. And if not, what modifications need to be done, or conditions to be fulfilled, to ensure that ELMs are practically freed of SB.

This work investigates if baseline ELMs demonstrate any characteristics of Spectral Bias. Comparisons are made against ANNs investigated in [3] which had clearly demonstrated SB. Investigations show that ELMs are less prone to SB, but in specific cases SB continues to exist. This work does not seek to justify these observations from a theoretical viewpoint. Instead, it evaluates means it of mitigation, and comes up with mechanisms for complete elimination of Spectral Bias from ELMs. This brings into the realm of feasibility the application of ELMs in variants of PINNs for resolving governing equations of physical systems especially where higher frequencies are involved, possibly opening the path for a plethora of developments and applications in diverse domains.

The rest of this paper is organized as follows. Section II provides the linkage between NTK theory, Spectral Bias and demonstrates SB on different simple but relevant functions. Section III explains the basic formulation of Extreme Learning Machines. Section IV investigates ELMs for SB on the same functions as used on ANNs, and then discusses and demonstrates mechanisms by which SB can be eliminated. Conclusions are drawn in Section V.

## II. SPECTRAL BIAS FROM NTK THEORY AND OBSERVATIONS ON ANNs

This section first expresses the fundamental result from Neural Tangent Kernel Theory [1, 11] and from there deduces the varying convergence rates of different frequency components of the function being trained, for a conventional MSE-loss based ANN.

Let $f(x,\theta)$ represent a scalar valued fully connected ANN with weights $\theta$ initialized by a Gaussian distribution. Considering a training data set $\{X_{trn}, Y_{trn}\}$ composed of N samples, one may express inputs $X_{trn}$ as $(x_i)_{i=1}^N$ and the corresponding outputs $Y_{trn}$ as $(y_i)_{i=1}^N$. If the ANN is trained using the Mean Square Error loss function

$$\mathbf{L}(\theta) = \frac{1}{N}\sum_{i=1}^{N}\left(f(x_i,\theta)-y_i\right)^2 \tag{1}$$

with a very small value of learning rate parameter η, then, using the derivation of Jacot et al [1, 11], one may define the Neural Tangent Kernel (NTK) operator $K$, with entries given by

$$K_{i,j} = K(x_i, x_j) = \left\langle \frac{\partial f(x_i,\theta)}{\partial \theta}, \frac{\partial f(x_j,\theta)}{\partial \theta} \right\rangle. \tag{2}$$

The NTK theory shows that, under the above conditions and using a gradient descent approach to training, the kernel $K$ converges to a deterministic value and does not change even if the width of the network hidden layer/s increase towards infinity.

Further, it can be shown that [12]

$$\frac{df(X_{trn},\theta(t))}{dt} \approx -K \cdot \left(f(X_{trn},\theta(t))-Y_{trn}\right) \tag{3}$$

where $\theta(t)$ denotes the parameters of the network at iteration $t$; the vector form of differential equation (3) may be observed. Solution of (3) may be expressed as

$$f(X_{trn}, \theta(t)) \approx (I - e^{-Kt})Y_{trn} \quad (4)$$

The kernel $K$ being square symmetric and positive semi-definite, we can express its spectral decomposition as

$$K = Q\Lambda Q^T \quad (5)$$

where $Q$ is an orthogonal matrix with $i^{th}$ column as the eigenvector $q_i$ of $K$, and $\Lambda$ is a diagonal matrix with entries $\lambda_i$ as the corresponding eigenvalues. Also note that $Q^T = Q^{-1}$, and as

$$e^{-Kt} = Qe^{-\Lambda t}Q^T \quad (6)$$

From (4), one may write

$$\left(f(X_{trn}, \theta(t)) - Y_{trn}\right) \approx -e^{-Kt}Y_{trn} \quad (7)$$

where and on substituting from (6), (7) yields

$$\left(f(X_{trn}, \theta(t)) - Y_{trn}\right) \approx -Qe^{-\Lambda t}Q^T Y_{trn}$$

which can be further written as

$$Q^T\left(f(X_{trn}, \theta(t)) - Y_{trn}\right) \approx -e^{-\Lambda t}Q^T Y_{trn} \quad (8)$$

Equation (8) can be written in expanded form as shown below. Eq. (9) shows that the $i^{th}$ component of the absolute error, $\left|q_i^T \cdot (f(X_{trn}, \theta(t)) - Y_{trn})\right|$, will decay approximately exponentially at the rate $\lambda_i$. That is, components of the target function that correspond to kernel eigenvectors with larger eigenvalues, will be learnt faster. The larger eigenvalues correspond to the larger spectral wavelengths and hence the smaller (lower) frequencies, and vice-versa, see e.g. [13] and [14]. Thus, for a fully connected ANN with MSE loss function and small learning rate parameter, in the process of training *the lower frequency components of the target function learn faster than the higher ones*.

$$\begin{bmatrix} q_1^T \\ q_2^T \\ q_3^T \\ . \\ . \\ . \\ q_N^T \end{bmatrix} \left(f(X_{trn}, \theta(t)) - Y_{trn}\right) = \begin{bmatrix} e^{-\lambda_1 t} & & & & \\ & e^{-\lambda_2 t} & & & \\ & & e^{-\lambda_3 t} & & \\ & & & . & \\ & & & & . \\ & & & & & e^{-\lambda_N t} \end{bmatrix} \begin{bmatrix} q_1^T \\ q_2^T \\ q_3^T \\ . \\ . \\ . \\ q_N^T \end{bmatrix} Y_{trn} \quad\quad \ldots\ldots \quad (9)$$

To validate the derivations shown above, a small set of numerical experiments were performed using some simple equations incorporating varying frequencies, and the results are shown below. It may be noted that these same equations are used later in Sec. IV to investigate the persistence of Spectral Bias in ELMs. There are basically two equations, one with combined frequency terms in the RHS (right hand side), and the other with a single frequency term at RHS, but with varying frequencies.

The combined frequency equation is:

$$f(x) = \sum_{k=1}^{5} \sin(2kx)/(2k), \quad x \in [-\pi, \pi] \quad (10)$$

the boundary conditions are set at $f(x)=0$ at both ends of the given range.

While the single frequency equation is:

$$f(x) = -\frac{1}{k^2}\sin kx, \text{ for k = 2, 6 and 10, } x \in [-\pi, \pi] \quad (11)$$

Equation (10) is solved on a fully-connected feedforward ANN in supervised learning mode, MSE cost function, with training data extracted from the closed form. The considered ANN has two hidden layers with 100 neurons per hidden layer, with tanh() as the activation function. While the typical plots of convergence and accuracy against the number of iterations are made, it is considered more pertinent here to show the plots of the developing solution (red curve) against the closed form solution, in blue. The plots are made for every 1000 (referred as 1 K) iterations till convergence, and only the plots at 1 K and 5 K iterations are shown, in figs. 1 and 2.

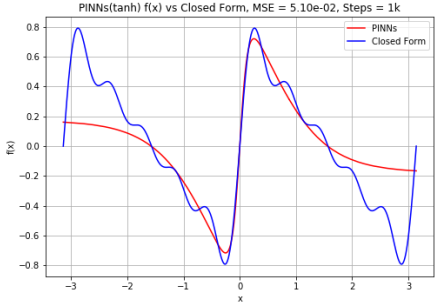
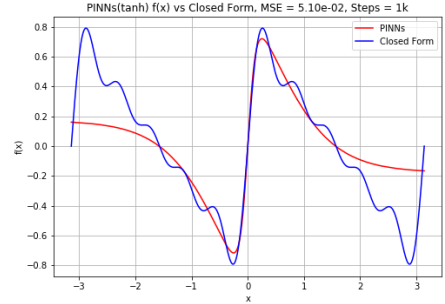

Fig. 1. Developing solution for $f(x)$, eq. 10, at 1 K iterations (red) against the closed form solution (blue).

Fig. 2. Developing solution for $f(x)$, eq. 10, at 5 K iterations (red) against the closed form solution (blue).

It can be seen clearly from figs. 1 & 2 that while the ANN has quickly captured the low frequency components of the function, it has not been able to resolve the relatively higher frequency components (the wiggles). This is exactly according to the prior discussions. In fact, the function is fully captured only at 18 K iterations.

Table 1. below summarizes the results from an analogous series of runs made on data generated from eq. (11), at different frequency values k. It clearly shows that as the frequency increases, convergence becomes more difficult and longer to attain. Again, this is consistent with the prior discussions in this section.

Table 1. Convergence of function of single sinusoids at different frequencies; activation function tanh(.)

| Frequency, k | Number of iterations at which convergence is achieved for the function |
|---|---|
|  | $f(x)$, (eq. 11), for conventional ANN |
| 2 | 440 |
| 6 | 2000 |
| 10 | 3600 |

### III. BASIC EXTREME LEARNING MACHINE APPROACH

Architecturally, the Extreme Learning Machine is a Single Hidden Layer Feedforward Network with the characteristic that among the two weight layers, the first, i.e., between input and hidden node layers, the weights are selected randomly leaving only the second weight layer to be solved for as a function of data characteristics and the selected first layer weights. This obviates the need for trans-layer iterative backpropagation as typical of feedforward ANNs. Instead, the solution for the second weight layer is obtained directly by matrix inversion following the approach of Huang et al [8] summarized below.

Let there be N distinct data samples $(\mathbf{x}_j, \mathbf{t}_j)$, for $j = 1, .., N$ where each $\mathbf{x}_j$ is a vector of $n$ components $x_{ij}$, $i=1, ..., n$ and each $\mathbf{t}_j$ a vector of m components $t_{kj}$, $k = 1,..., m$. Let the number of nodes in the hidden layer be $L$, and each node be denoted by index $l = 1, ..., L$.

The induced local field (i.e. influence of previous layer nodes) at node $l$ of the hidden layer for a data sample $j$ can be expressed as

$$v_{lj} = \sum_{i=1}^{n} w_{li} x_{ij} + b_l \qquad (12)$$

which $b_l$ is the bias value for node $l$. Now if $g(.)$ is the activation function at each hidden layer node, then the output from a hidden layer node is simply

$$y_{lj} = g(v_{lj}) \qquad (13)$$

The induced local field at node $k$ of the output layer for a data sample j can be expressed as

$$z_{kj} = \sum_{l=1}^{L} \beta_{kl} y_{lj} \qquad (14)$$

note that there are no biases and also no activation function for the hidden layer nodes, so $z_{kj}$ is also the final output for sample $j$ from the node $k$. We have used $\beta_{kl}$ to denote the connecting weight from hidden layer node $l$ to output layer node $k$.

Now comes the crucial step that differentiates the ELM from the ANN formalism: we want to force the output value $z_{kj}$ to equal the targeted value $t_{kj}$, i.e., in principle

$$\sum_{j=1}^{N} \sum_{k=1}^{m} |z_{kj} - t_{kj}| = 0 \qquad (15)$$

It can be shown that (15) will hold true only under the conditions $z_{kj} = t_{kj}$, for $\forall j, k$ *as defined above* (16)

Substituting (16) in (14), one obtains

$$\sum_{l=1}^{L} \beta_{kl} y_{lj} = t_{kj} \qquad (17)$$

Eq. (17) is a matrix equation, and applying the transpose operation to either side, one may write

$$\left( \sum_{l=1}^{L} \beta_{kl} y_{lj} \right)^T = t_{kj}^T \qquad (18)$$

which on application of basic associative rules on the sums and products of transposes of matrices, can be written as

$$\sum_{l=1}^{L} y_{jl} \beta_{lk} = t_{jk} \qquad (19)$$

Recalling that $l$ is the index for hidden layer nodes varying from *1 to L*, and $k$ for output nodes varying from *1 to m*, one may alternately represent (19) in the form of a matrix equation

$$\begin{array}{cc} [Y_{jl}] & [\beta_{lk}] = [T_{jk}] \\ (N \times L) & (L \times m) \quad (N \times m) \end{array} \qquad (20)$$

Eq. (20) is used to solve for the matrix $[\beta_{lk}]$. The equation can be written in more compact form as

$$H\beta = T \qquad (20a)$$

where it is obvious $H \equiv [Y_{jl}]$.

One may note that none of these matrices are square. The approach to the solution of (20a) is to express it in the form

$$\hat{\beta} = H^\dagger T \qquad (21)$$

where $H^\dagger$ is the generalized Moore-Penrose inverse of $H$.

## IV. NUMERICAL EXPERIMENTATION AND RESULTS

Numerical investigations aim to evaluate the performance of ELMs on the multi-frequency case of eq. (10), and the different single-frequency cases of eq. (11), extract information on spectral bias and establish conditions for its mitigation and possible elimination. Performance of ELMs depends on three hyperparameters:
- the number of nodes in the hidden layer, denoted L
- the activation function used in the hidden layer nodes
- the characteristics of the random weight matrix between the input and hidden layers.

In almost all cases the input range of $[-\pi, +\pi]$ is discretized into a total of 1000 points among which 800 (4 out of every 5 over the range) are taken for training and the rest for test. With reference to sec. III, this implies N is 800. Further, the number of hidden layer nodes L is taken as 800, but experiments performed with varying values like 400 and 1600 (and other values not reported here). All the cases have one input, i.e. $x$, and one output $f(x)$, implying n = m = 1. The activation function used here is tanh(.), based on our past experiences with ANNs and ELMs, this seems to give the best performance.

The characteristics of the random weight matrix are the major subject of variability and concomitant investigations. First, the need for repeatability as typical of Machine Learning investigations, engenders the need to use random number seeds so that the same first-layer-matrix can be regenerated for given cases. Libraries in Python allow random number (RN henceforth) generation with seed, but the resultant RNs follow a uniform distribution within some range of values. So, this is the first possible characteristic of RN matrix, that is, RN with uniform distribution.

Further investigations are performed on RN matrices where the weights follow a normal (Gaussian) distribution. By default, this takes mean zero and Standard Deviation (SD) as 1. At the next level, numerical experiments are performed with SD set at different values. It may also be noted that randomization of the bias vector for a case follows exactly the approach taken for the RN matrix, for that case. The results of all these investigations are presented below.

First, the performance of ELMs on eq. (10) are shown in figs. 3 and 4. Cases are for test data, which is also true for all succeeding figures. Fig. 3 shows results for RN generated using uniform distribution, while fig. 4 shows results for normal distribution with default SD (=1). Two important aspects are immediately observable: one, that the distribution of the RNs in the first matrix have a profound impact and normal distribution definitely performs better for *higher frequencies* compared to uniform distribution, and two, the one shot solution for ELM with normal RNs *achieves the accuracy level obtained on ANN with 18 K iterations*.

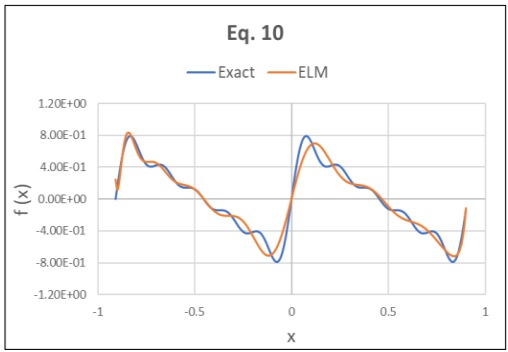

Fig. 3. Solution for *f(x)*, eq. 10, from ELM with RN matrix from uniform distribution, against the closed form solution (blue).

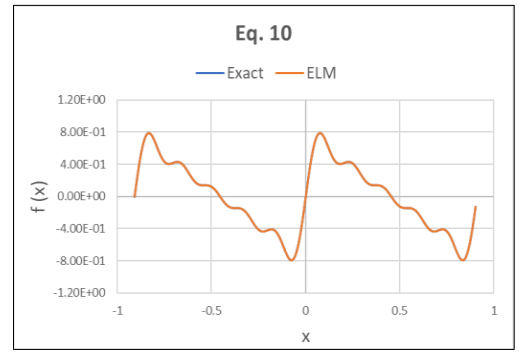

Fig. 4. Solution for *f(x)*, eq. 10, from ELM with RN matrix from normal distribution, against the closed form solution (blue).

Figures 5-7 show results for eq. (11), with k set at 2, 6 and 10 respectively, using uniform RN distribution. While the lower frequencies are captured properly, it is clear that the higher frequency of k = 10 is not reproduced. This result is important, as this shows, prima-facie, that *ELMs are not completely free of Spectral Bias*.

Next, figs. 8-11 illustrate results on eq. (11), with k set at 2, 6, 10 and 20. Importantly, the RN weights are obtained from default normal distribution. It is observed that frequencies 2, 6 and 10 are captured well, but the highest frequency of 20 could not be resolved. So *even with normally distributed RNs in the first weight matrix, Spectral Bias is seen to persist*. As mentioned in Sec. I, the authors here do not attempt any theoretical justification for the persistence of SB, but instead, aim to work at mitigation / elimination of SB from ELMs.

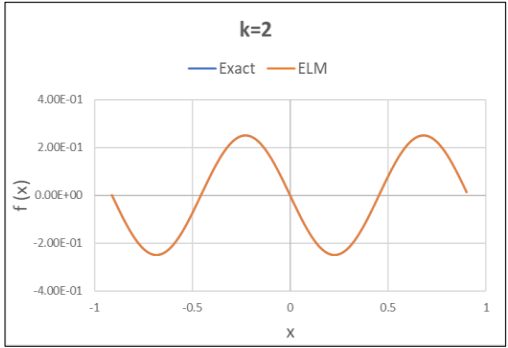

Fig. 5. Solution for $f(x)$, eq. 11 with k = 2, from ELM with RN matrix from uniform distribution, against the closed form solution (blue).

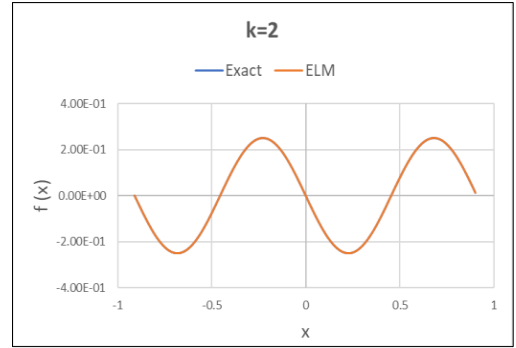

Fig. 8. Solution for $f(x)$, eq. 11 with k = 2, from ELM with RN matrix from normal distribution, against the closed form solution (blue).

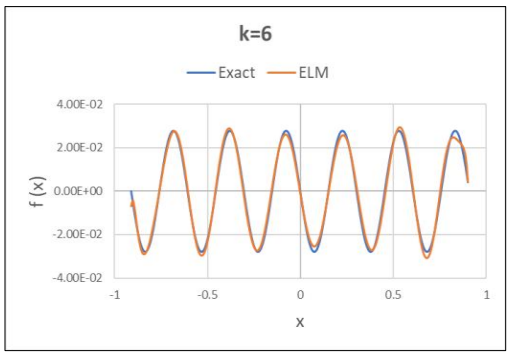

Fig. 6. Solution for $f(x)$, eq. 11 with k = 6, from ELM with RN matrix from uniform distribution, against the closed form solution (blue).

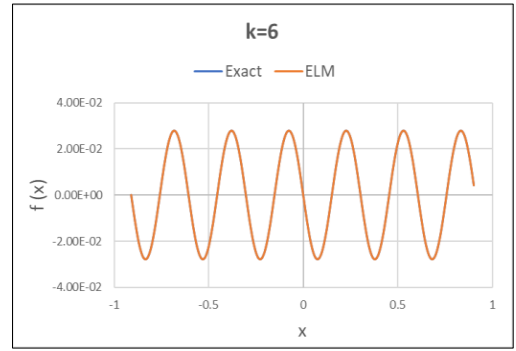

Fig. 9. Solution for $f(x)$, eq. 11 with k = 6, from ELM with RN matrix from normal distribution, against the closed form solution (blue).

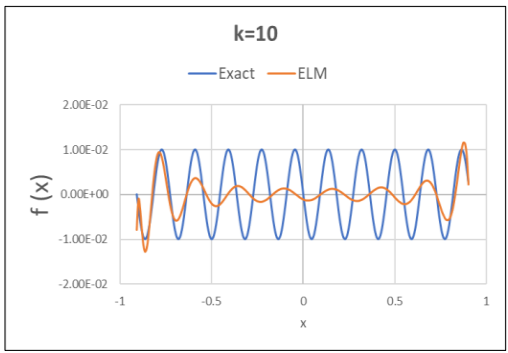

Fig. 7. Solution for $f(x)$, eq. 11 with k = 10, from ELM with RN matrix from uniform distribution, against the closed form solution (blue).

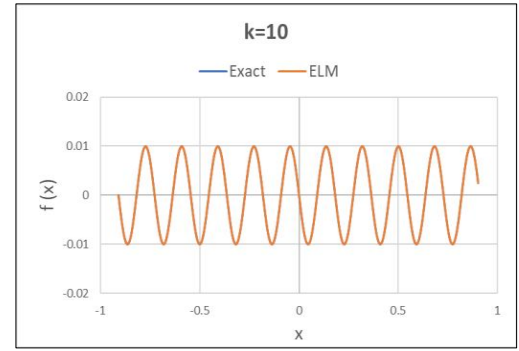

Fig. 10. Solution for $f(x)$, eq. 11 with k = 10, from ELM with RN matrix from normal distribution, against the closed form solution (blue).

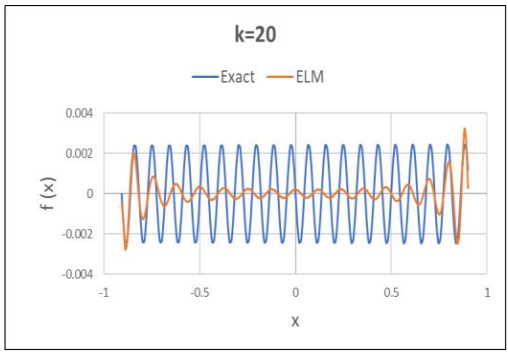 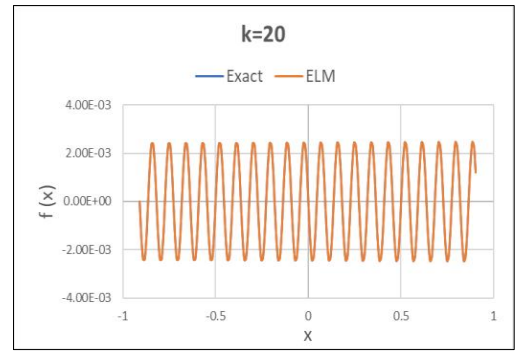

Fig. 11. Solution for *f(x)*, eq. 11 with k = 20, from ELM with RN matrix from normal distribution, against the closed form solution (blue).

Fig. 12. Solution for *f(x)*, eq. 11 with k = 20, from ELM with RN matrix from normal distribution *and SD forced to 20*, against the closed form solution (blue).

From the observation in fig. 11 that ELMs continue to exhibit SB even if the first layer matrix is drawn from a Gaussian distribution, it is considered pertinent to refer to the concept of Fourier Feature Embeddings [2][15], where the SD of the distribution of the random weight matrix between the input and first hidden layer is taken to equal the strongest frequency component of the function under consideration. Implicit in the above statement is the fact that the first weight matrix in such ANNs is not obtained by iterative solution but in advance from a random distribution. It has been demonstrated, see [15], that choice of such Fourier Feature Embeddings indeed significantly mitigates SB in the considered ANN.

The above concept is investigated in the case of k = 20, where the SD of the normally distributed RNs of the first weight matrix is taken equal to 20. As observed in fig. 12, this facilitates capture of the function with high accuracy. The implication is that implementation of the concept of Fourier Feature Embeddings can indeed eliminate persisting Spectral Bias in ELMs.

Next, results are presented from runs made on a much higher frequency of 50, in figs. 13-17. Fig. 13 shows results for RNs of the first weight matrix generated from default normal distribution (SD = 1). This is expected, as the same conditions failed to generate correct results even for a lower frequency of k = 20. Carrying on from the concept of Fourier Feature Embeddings, the SD of the first matrix is forced to a value of 50. Now even the high frequency of 50 is captured accurately, as shown in fig. 14. This somewhat reinforces the application of Fourier Feature Embeddings to mitigate persisting Spectral Bias in ELMs.

The next figure, fig. 15, seems to overturn this view. Here the RNs of the first matrix are created with SD set at 20. And surprisingly, though this is well below the function's dominant frequency of 50, the function is accurately reproduced by the ELM. This important result leads to the perception that, at some intermediate value of SD between 1 and 20, there is an inflexion point where the RNs of the first matrix transit from "inability" to capture the function to its accurate representation. Figs. 16 with SD = 7, and 17 with SD = 5, provide this transition zone. The observations are self-explanatory.

From the above observations one may deduce that unlike in the case of ANNs where Spectral Bias applies precisely and application of Fourier Feature Embeddings can indeed mitigate the same, in ELMs Spectral Bias is also observed at higher frequencies but if the Standard Deviation of the RNs in the first matrix are chosen at or even near the dominant frequency of the function, the SB is eliminated.

At this point one is in a position to formulate a strategy for elimination of Spectral Bias from ELMs. The core idea follows from the two facts that, one, Spectral Bias observed at the highest frequencies of the represented function gets eliminated when the natural choice of SD for the RNs of the first weight matrix is at or even close to these frequencies, and two, setting the SD for the RNs of the first matrix at close to the highest frequencies naturally prevents formation of SB at the lower frequencies composing the function. *Then all that one has to do to eliminate Spectral Bias from Extreme Learning Machines is to set the Standard Deviation of the normally distributed random numbers of the first matrix at levels close to the highest frequencies of the function under consideration.*

Just to further verify this point, fig. 18 presents result for a case of k = 6, with SD set at 50. This helps to reconfirm the second fact stated in the prior paragraph.

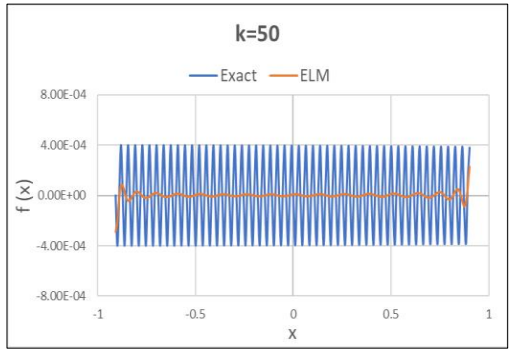

Fig. 13. Solution for *f(x)*, eq. 11 with k = 50, from ELM with RN matrix from normal distribution, against the closed form solution (blue).

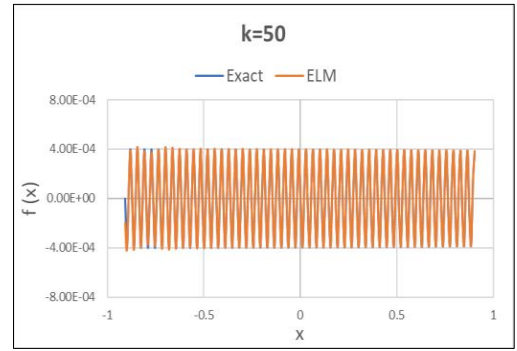

Fig. 16. Solution for *f(x)*, eq. 11 with k = **50**, from ELM with RN matrix from normal distribution *and SD forced to **7***, against the closed form solution (blue).

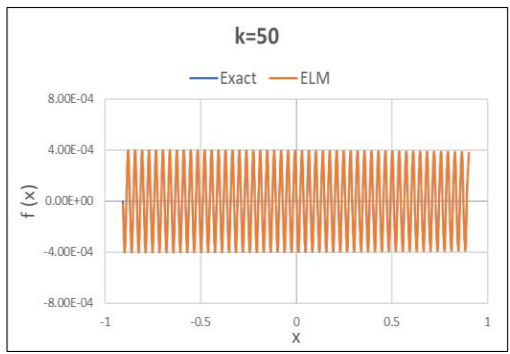

Fig. 14. Solution for *f(x)*, eq. 11 with k = 50, from ELM with RN matrix from normal distribution *and SD forced to 50*, against the closed form solution (blue).

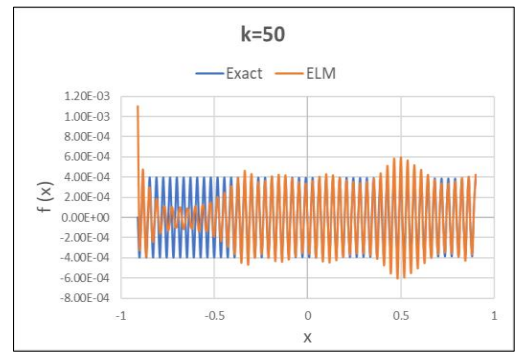

Fig. 17. Solution for *f(x)*, eq. 11 with k = **50**, from ELM with RN matrix from normal distribution *and SD forced to **5***, against the closed form solution (blue).

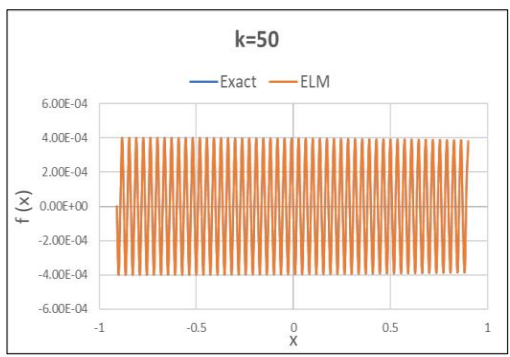

Fig. 15. Solution for *f(x)*, eq. 11 with k = **50**, from ELM with RN matrix from normal distribution *and SD forced to **20***, against the closed form solution (blue).

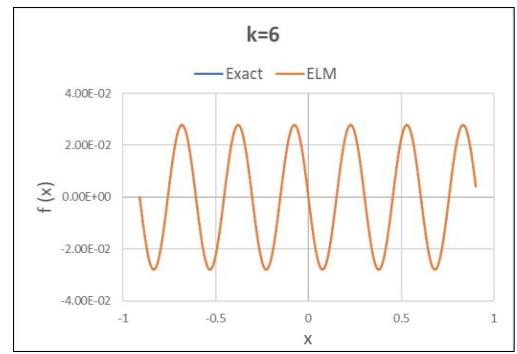

Fig. 18. Solution for *f(x)*, eq. 11 with k = **6**, from ELM with RN matrix from normal distribution *and SD forced to **50***, against the closed form solution (blue).

All the results presented in this paper have been run with a value of L, i.e., number of nodes in the hidden layer, set at 800. However, the authors have run some cases, including those reflecting inadequacy of representation of a considered function, at values of L set at 400 and 1600. No change in results have been observed. All computations are performed on Google Colaboratory using CPUs alone, where the solution times for L set at 400, 800 and 1600 are 42, 43 and 58 milliseconds respectively.

## V. CONCLUSIONS

Spectral Bias that is observed consistently in feedforward Artificial Neural Networks, is detected occasionally in Extreme Learning Machines.

The mechanisms of operation of Extreme Learning Machines make them naturally amenable to implementation of variants of Fourier Feature Embeddings, which is seen to eliminate Spectral Bias from ELMs across all frequencies composing the function under consideration.

This brings into the domain of feasibility the application of Extreme Learning Machines to problems where resolution of high frequency components of functions is considered essential, like in the case of Physics Informed Neural Networks.

## ACKNOWLEDGMENTS

This work was partially supported under Aeronautics Research and Development Board, Government of India, Aerodynamics Panel Grant No. 2051. The authors also acknowledge support provided by all coworkers, included past undergraduate students of Mahindra University, to this work.

## REFERENCES

[1] Arthur Jacot, Franck Gabriel Clement Hongler, "Neural tangent kernel: Convergence and generalization in neural networks", Advances in Neural Information Processing Systems, 2018, pp.8571-8580.
[2] Matthew Tancik et al, "Fourier features let networks learn high frequency functions in low dimensional domains", arXiv:2006.10739, also NeurIPS Proceedings, 2020.
[3] Mayank Deshpande et al, "Investigations on convergence behaviour of Physics Informed Neural Networks across spectral ranges and derivative orders", arXiv:2301.02790, also, *2022 IEEE Symposium Series on Computational Intelligence* (SSCI), Singapore, 2022, pp. 1172-1179, doi: 10.1109/SSCI51031.2022.10022020.
[4] G.G. Stokes, "On some cases of fluid motion", Transactions of the Cambridge Philosophical Society, 8, 1843, pp. 105–137.
[5] J.C. Maxwell, "A dynamical theory of the electromagnetic field", Philosophical Transactions of the Royal Society, 165, 1865, pp. 459–512.
[6] M. Raissi, P. Perdikaris, and G.E. Karniadakis, "Physics-informed neural networks: A deep learning framework for solving forward and inverse problems involving nonlinear partial differential equations", Journal of Computational Physics, 378, 2019, pp. 686-707.
[7] Lawal, Z.K., Yassin, H., Lai, D.T.C. and Che Idris, A., 2022. Physics-Informed Neural Network (PINN) Evolution and Beyond: A Systematic Literature Review and Bibliometric Analysis. Big Data and Cognitive Computing, 6(4), p.140.
[8] G.B. Huang, Q.Y. Zhu and C.K. Siew, "Extreme Learning machine: Theory and applications," Neurocomputing, vol. 70, no. 1–3, Dec. 2006, pp. 489–501.
[9] Guang-Bin Huang et al, "Extreme Learning Machine for Regression and Multiclass Classification," IEEE Transaction on Systems, Man and Cybernetics-Part B: Cybernetics, Vol. 42, No.2, Apr 2012, pp. 513-529.
[10] R. R. Annapureddy, A. K. Bhattacharya and Niranjan Reddy, "Adaptive Critic Design for Extreme Learning Machines applied to noisy and drifting industrial processes," *2018 IEEE Symposium Series on Computational Intelligence* (SSCI), Bangalore, India, 2018, pp. 327-334, doi: 10.1109/SSCI.2018.8628664.
[11] Sanjeev Arora et al, "On exact computation with an infinitely wide neural net", Advances in Neural Information Processing Systems, 2019, pp. 8141–8150.
[12] Jaehoon Lee et al, "Wide neural networks of any depth evolve as linear models under gradient descent", Advances in Neural Information Processing Systems, 2019, pp. 8572–8583.
[13] https://www.sciencedirect.com/topics/mathematics/smallest-eigenvalue, Sec. 4.3, para 2; last accessed Apr 8, 2023.
[14] Sandip Mazumder, *Numerical Methods for Partial Differential Equations: Finite Difference and Finite Volume Methods*, Chap. 4, ISBN: 978-0128498941, 2016.
[15] S.Wang, H. Wang and P. Perdikaris, "On the eigenvector bias of Fourier Feature Networks: From Regression to solving multi-scale PDEs with Physics Informed Neural Networks", Computer Methods in Applied Mechanics and Engineering, Vol. 384, 2021.